\documentclass[]{ecai} 

\usepackage{latexsym}
\usepackage{amssymb}
\usepackage{amsmath}
\usepackage{amsthm}
\usepackage{booktabs}
\usepackage{enumitem}
\usepackage{graphicx}
\usepackage{xcolor}
\usepackage{subcaption}
\usepackage{multirow}
\usepackage{graphics}
\usepackage{mathtools}
\usepackage{stmaryrd}
\usepackage{adjustbox}
\SetSymbolFont{stmry}{bold}{U}{stmry}{m}{n}
\usepackage{tikz}
\usetikzlibrary{arrows,graphs,quotes,positioning}
\usepackage{anyfontsize}
\usepackage[hidelinks]{hyperref}
\usepackage{orcidlink}

\newtheorem{theorem}{Theorem}

\newtheorem{corollary}[theorem]{Corollary}
\newtheorem{proposition}[theorem]{Proposition}

\newtheorem{definition}{Definition}
\newtheorem{example}{Example}

\newcommand{\BibTeX}{B\kern-.05em{\sc i\kern-.025em b}\kern-.08em\TeX}

\newcommand{\DLogic}{\mathcal{SROI}^-}
\newcommand{\OurMethod}{AConE}
\newcommand{\ConceptNames}{\mathbf{C}}
\newcommand{\RoleNames}{\mathbf{R}}
\newcommand{\IndividualNames}{\mathbf{I}}
\newcommand{\ABox}{\mathcal{A}}
\newcommand{\TBox}{\mathcal{T}}
\newcommand{\RBox}{\mathcal{R}}
\newcommand{\KBase}{\mathcal{K}}
\newcommand{\Int}{\mathcal{I}}
\newcommand{\ConeAlgebra}{Cone Algebra}
\newcommand{\Real}{\mathbb{R}}
\newcommand{\Complex}{\mathbb{C}}
\newcommand{\Embedding}{\mathcal{E}}

\newcommand{\RoleAssertion}[3]{#1(#2, #3)}
\newcommand{\ConceptAssertion}[2]{#1(#2)}
\newcommand{\inverse}[1]{#1^{-}}
\newcommand{\nominal}[1]{\{#1\}}
\newcommand{\Disj}[2]{\mathsf{Disj}(#1, #2)}
\newcommand{\Trans}[1]{\mathsf{Trans}(#1)}
\newcommand{\Reflex}[1]{\mathsf{Ref}(#1)}
\newcommand{\Irref}[1]{\mathsf{Irref}(#1)}
\newcommand{\Symmetry}[1]{\mathsf{Sym}(#1)}
\newcommand{\Asymmetry}[1]{\mathsf{Asym}(#1)}

\newcommand{\ConceptName}[1]{\mathsf{#1}}
\newcommand{\Individual}[1]{\mathsf{#1}}
\newcommand{\Ronaldo}{\Individual{C.Ronaldo}}
\newcommand{\Messi}{\Individual{L.Messi}}

\newcommand{\Trashorras}{\Individual{R.Trashorras}}
\newcommand{\WorldCup}{\Individual{WorldCup}}
\newcommand{\EuroCup}{\Individual{EuroCup}}

\newcommand{\ArgentinaNFT}{\Individual{ArgentinaNFT}}

\newcommand{\Rosario}{\Individual{Rosario}}

\newcommand{\Relation}[1]{\mathsf{#1}}
\newcommand{\playsFor}{\Relation{playsFor}}
\newcommand{\teamMate}{\Relation{teamMate}}
\newcommand{\athleteWon}{\Relation{athleteWon}}
\newcommand{\teamWon}{\Relation{teamWon}}
\newcommand{\birthPlaceOf}{\Relation{birthPlaceOf}}
\newcommand{\bornAt}{\Relation{bornAt}}

\newcommand{\embedding}[1]{\boldsymbol{#1}}
\newcommand{\answers}[2]{\llbracket #1 \rrbracket_{#2}}
\newcommand{\Cone}{\mathsf{C}}
\newcommand{\MCone}{\mathsf{MC}}
\newcommand{\SetCones}{\mathcal{C}}
\newcommand{\EmptyCone}{\Cone_\bot}
\newcommand{\FullCone}{\Cone_\top}
\newcommand{\SingletonCone}[1]{\Cone_{#1}}
\newcommand{\ProperCone}[2]{\Cone_{{#1}\to{#2}}}
\newcommand{\Rot}{\mathsf{R}}
\newcommand{\Rotation}[3]{\Rot\langle #1, #2, #3 \rangle}

\newcommand{\FullMCone}{\MCone_\top}
\newcommand{\UperBound}{\embedding{h_U}}
\newcommand{\LowerBound}{\embedding{h_L}}
\newcommand{\ConeAxis}{\embedding{h_{\mathit{ax}}}}
\newcommand{\UperAngle}{\embedding{\theta_U}}
\newcommand{\LowerAngle}{\embedding{\theta_L}}
\newcommand{\AxAngle}{\embedding{\theta_{\mathit{ax}}}}
\newcommand{\ApAngle}{\embedding{\theta_{\mathit{ap}}}}
\newcommand{\EntityVector}{\embedding{h}^{*}}
\newcommand{\EntityAngle}{\embedding{\theta}^{*}}
\newcommand{\NegativeSample}[1]{\embedding{h_{#1}'}}
\newcommand{\PositiveSample}{\embedding{h^*}}
\newcommand{\QueryEmbedding}{\embedding{q}}

\newcommand{\GeoOp}[1]{\mathcal{P_{#1}}}
\newcommand{\GeoOpExists}{\GeoOp{\exists}}
\newcommand{\GeoOpConj}{\GeoOp{\sqcap}}
\newcommand{\GeoOpDisj}{\GeoOp{\sqcup}}
\newcommand{\GeoOpNeg}{\GeoOp{\neg}}

\newcommand{\rotation}{\embedding{r}}
\newcommand{\rotationL}{\embedding{r_L}}
\newcommand{\rotationU}{\embedding{r_U}}
\newcommand{\VecOne}{\mathbf{1}}
\newcommand{\AxRAngle}{\embedding{\embedding{\theta_{\mathit{ax},r}}}}
\newcommand{\ApRAngle}{\embedding{\embedding{\theta_{\mathit{ap},r}}}}

\newcommand{\VectorComponent}[2]{[#1]_{#2}}
\newcommand{\UperBoundJ}[1]{\VectorComponent{\UperBound}{#1}}
\newcommand{\LowerBoundJ}[1]{\VectorComponent{\LowerBound}{#1}}
\newcommand{\UperAngleJ}[1]{\VectorComponent{\UperAngle}{#1}}
\newcommand{\LowerAngleJ}[1]{\VectorComponent{\LowerAngle}{#1}}

\newcommand{\ApAngleJ}[1]{\VectorComponent{\ApAngle}{#1}}
\newcommand{\rotationLJ}[1]{\VectorComponent{\rotationL}{#1}}
\newcommand{\rotationUJ}[1]{\VectorComponent{\rotationU}{#1}}
\newcommand{\AxRAngleJ}[1]{\VectorComponent{\AxRAngle}{#1}}
\newcommand{\ApRAngleJ}[1]{\VectorComponent{\ApRAngle}{#1}}

\newcommand{\SemanticAverage}{\operatorname{SemanticAverage}}
\newcommand{\CardMin}{\operatorname{CardMin}}

\newcommand{\distance}[1]{\operatorname{d}_{\operatorname{#1}}}
\newcommand{\CombDistance}{\distance{comb}}
\newcommand{\InDistance}{\distance{i}}
\newcommand{\OutDistance}{\distance{o}}

\newcommand{\st}{:}

\begin{document}


\begin{frontmatter}


\paperid{2036} 


\title{Generating $\DLogic$ Ontologies via Knowledge Graph Query Embedding Learning}


\author[AB]{\fnms{Yunjie}~\snm{He}\thanks{Corresponding Author. Email: yunjie.he@ki.uni-stuttgart.de}~\orcidlink{0009-0005-4461-2863}}
\author[A]{\fnms{Daniel}~\snm{Hernandez}~\orcidlink{0000-0002-7896-0875}}
\author[A]{\fnms{Mojtaba}~\snm{Nayyeri}~\orcidlink{0000-0002-9177-0312}} 
\author[A]{\fnms{Bo}~\snm{Xiong}~\orcidlink{0000-0002-5859-1961}}
\author[AB]{\fnms{Yuqicheng}~\snm{Zhu}~\orcidlink{0000-0001-5845-5401}}
\author[BD]{\fnms{Evgeny}~\snm{Kharlamov}~\orcidlink{0000-0003-3247-4166}}
\author[AC]{\fnms{Steffen}~\snm{Staab}~\orcidlink{0000-0002-0780-4154}}

\address[A]{University of Stuttgart}
\address[B]{Bosch Center for Artificial Intelligence}
\address[C]{University of Southampton}
\address[D]{University of Oslo}

\begin{abstract}
Query embedding approaches answer complex logical queries over incomplete knowledge graphs (KGs) by computing and operating on low-dimensional vector representations of entities, relations, and queries. However, current query embedding models heavily rely on excessively parameterized neural networks and cannot explain the knowledge learned from the graph. We propose a novel query embedding method, {\OurMethod}, which explains the knowledge learned from the graph in the form of $\DLogic$ description logic axioms while being more parameter-efficient than most existing approaches. {\OurMethod} associates queries to $\DLogic$ description logic concepts. Every $\DLogic$ concept is embedded as a cone in complex vector space, and each $\DLogic$ relation is embedded as a transformation that rotates and scales cones.
We show theoretically that {\OurMethod} can learn $\DLogic$ axioms, and defines an algebra whose operations correspond one-to-one to $\DLogic$ description logic concept constructs. Our empirical study on multiple query datasets shows that {\OurMethod} achieves superior results over previous baselines with fewer parameters. Notably on the WN18RR dataset, {\OurMethod} achieves significant improvement over baseline models. We provide comprehensive analyses showing that the capability to represent axioms positively impacts the results of query answering.
\end{abstract}

\end{frontmatter}


\section{Introduction}

\begin{figure}[t]
 \vspace{-5pt}
  \begin{center}
    \resizebox{\hsize}{!}{
    \begin{tikzpicture}
      \tikzset{kgNode/.style={draw, rounded corners, minimum height=1.7em, font=\scriptsize}}
      \tikzset{kgLabel/.style={rounded corners, fill=white, font=\scriptsize}}
      \tikzset{kgEdgeA/.style={draw, ->}}
      \tikzset{kgEdgeB/.style={dotted, ->}}
      \node[kgNode] (Ronaldo) at (-1,1) {$\Ronaldo$};
      \node[kgNode] (WorldCup) at (-3,2.75) {$\WorldCup$};
      \node[kgNode] (Messi) at (-5,4.5) {$\Messi$};
      \node[kgNode] (Thrashorras) at (-1,4.5) {$\Trashorras$};
      \node[kgNode] (ArgentinaNFT) at (-5,1) {$\ArgentinaNFT$};
      \node[kgNode] (Rosario) at (-8.5,4.5) {$\Rosario$};
      \path[kgEdgeB] (Messi) [bend left=40] edge node [kgLabel, midway] {$\athleteWon$} (WorldCup);
      \path[kgEdgeA] (ArgentinaNFT) [bend left=-40] edge node [kgLabel, midway] {$\teamWon$} (WorldCup);
      \path[kgEdgeA] (Messi) edge node [kgLabel, midway] {$\playsFor$} (ArgentinaNFT);
      \path[kgEdgeA] (Messi) [bend left=5] edge node [kgLabel, midway] {$\teamMate$} (Thrashorras);
      \path[kgEdgeA] (Thrashorras) [bend left=-20] edge node [kgLabel, midway] {$\teamMate$} (Messi);
      \path[kgEdgeA] (Messi) [bend left=-20] edge node [kgLabel, midway] {$\bornAt$} (Rosario);
      \path[kgEdgeB] (Rosario) [bend left=-20] edge node [kgLabel, midway] {$\birthPlaceOf$} (Messi);
      \path[kgEdgeB] (Ronaldo) [bend left=30] edge node [kgLabel, near start] {$\teamMate$} (Thrashorras);
      \path[kgEdgeA] (Thrashorras) [bend left=30] edge node [kgLabel, near start] {$\teamMate$} (Ronaldo);
    \end{tikzpicture}
    }
  \end{center}
  \caption{An example of an incomplete KG. The nodes represent entities, the edges with solid lines represent known atomic statements, whereas the edges with dotted lines represent missing atomic statements that must be inferred.}
  \label{Context_KG}
  \vspace{15pt}
\end{figure}
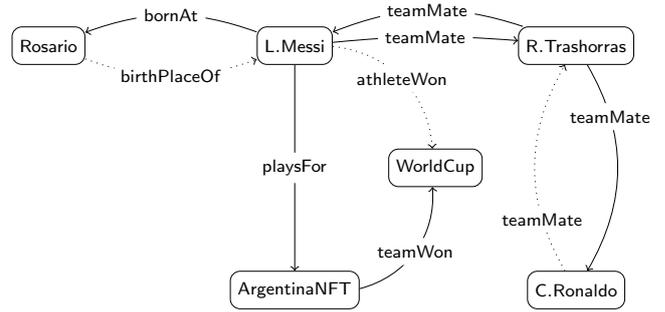

Knowledge Graphs (KGs) such as Wikidata \cite{Wikidata}, Freebase \cite{BollackerEPST08}, and YAGO \cite{SuchanekKW07} represent real-world facts as sets of \emph{triples} of the form $(s,p,o)$ which encode atomic assertions as $p(s,o)$. Graph database engines can store and query KGs efficiently using query languages such as SPARQL \cite{2013sparql} that can express a variety of queries that result of combining atomic queries called \emph{triple patterns}.
The first step of querying KGs is answering triple patterns with the stored triples. However, when querying incomplete Knowledge Graphs (KGs), some triples are not explicitly available in the triple store. As a result, these triples are neither included in the answers nor in the intermediate results. To provide plausible answers beyond what is known, these missing triples must be inferred. Figure \ref{Context_KG} shows an example, where the available triples $(\Messi, \playsFor, \ArgentinaNFT)$, and $(\ArgentinaNFT, \teamWon, \WorldCup)$
should suggest that triple $(\Messi, \athleteWon, \WorldCup)$ is a missing triple.
KG embedding methods can predict these missing triples \cite{TransE,DistMult,Rescal} by learning how to embed entities and relations into vector representations, which can be points or more complex geometric objects. These methods use these embeddings to answer triple patterns by computing plausibility scores by applying geometric operations. 

Query embedding methods \cite{querytobox,ConE,BetaE} go beyond querying triple patterns. They provide plausible answers to queries that combine triple patterns into first-order logic queries with logical connectives (e.g., negation ($\neg$), conjunction ($\wedge$), and disjunction ($\vee$)). However, current query embedding approaches are restricted to queries with a single unquantified variable and are called \emph{tree-form queries} because their \emph{computation graph} is a tree~\cite{CLQA_Survey}. The tree-form queries correspond to the $\DLogic$ description logic concepts that do not include the symbols $\top$ (the concept for all entities) nor $\bot$ (the concept for no elements), nor concept names (e.g., $\ConceptName{Athlete}$ or $\ConceptName{Team}$), but nominals (i.e., concepts with a unique element like $\nominal{\Ronaldo}$).

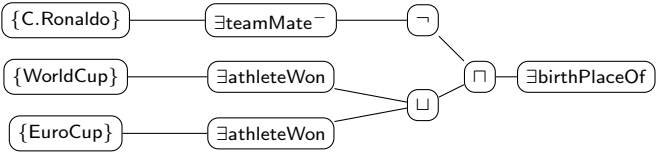
\begin{figure}
  \begin{center}
  \begin{tikzpicture}[%
      grow=180,%
      level distance=1.4cm,
      nodes={draw, rounded corners, font=\scriptsize, minimum height=1.3em},%
      edges from parent/.style={draw}]
    \node {$\exists\birthPlaceOf$}
    child {
      node {$\sqcap$}
      child [level distance=0.75cm] {
        node {$\neg$}
        child[level distance=2cm] {
          node {$\exists\inverse{\teamMate}$}
          child[level distance=2.7cm] {
            node {$\nominal{\Ronaldo}$}
          }
        }
      }
      child[sibling distance=0.75cm, level distance=0.75cm] {
        node {$\sqcup$}
        child[level distance=2cm] {
          node {$\exists\athleteWon$}
          child[level distance=2.7cm] {
            node {$\nominal{\WorldCup}$}
          }
        }
        child[sibling distance=0.75cm, level distance=2cm] {
          node {$\exists\athleteWon$}
          child[level distance=2.7cm] {
            node {$\nominal{\EuroCup}$}
          }
        }
      }
    };
  \end{tikzpicture}
  \end{center}
  \caption{Computation graph for the concept $C$ in Example~\ref{ex:running-example}.}%
  \label{fig:tree-form-concepts}
  \vspace{2em} 
\end{figure}

\begin{example}\label{ex:running-example}
The query seeking the birthplaces of the athletes who have won either the World Cup or the Europe Cup but do not play in the same team as $\Ronaldo$ can be expressed as the $\DLogic$ concept
\[
  C \equiv \begin{aligned}[t]
    \exists\birthPlaceOf.(&\neg(\exists\inverse{\teamMate}.\nominal{\Ronaldo})\; \sqcap\\[-2pt]
    &(\begin{aligned}[t]
        &\exists\athleteWon.\nominal{\WorldCup}\; \sqcup \\
        &\exists\athleteWon.\nominal{\EuroCup})),
      \end{aligned}
  \end{aligned}
\]
whose computation graph is depicted in Figure~\ref{fig:tree-form-concepts}.
\end{example}

Query embedding methods \cite{querytobox,ConE,BetaE} learn a model $M$ to predict the answers to queries by embedding entities and relations as geometrical objects of a vector space. These geometrical objects are used to recursively compute geometrical objects for all nodes in the computation graph of the query. Finally, the query embeddings are compared with the entity embeddings of the candidate answers to the query by using a similarity function that represents the plausibility of an answer.

The quality of query embedding methods depends on their ability to represent the \emph{logical patterns} satisfied by a knowledge graph. For example, some relations are symmetric (e.g., $\teamMate$) while others are antisymmetric (e.g., $\playsFor$); some relations are the inverse of other relations (e.g., $\bornAt$ and $\birthPlaceOf$) and some relations may be composed by others (e.g., if an athlete $x$ $\playsFor$ a team $y$ and $y$ $\teamWon$ a cup $z$, then $x$ $\athleteWon$ $z$). As these logical patterns influence the interplay of entities and relations, several works have studied their effect on KG embeddings \cite{RotatE,nayyeri2021losspattern,DBLP:conf/kdd/XiongZNXP0S22} and demonstrated that an embedding's ability to support them improves its link prediction quality. For query embeddings, similar support of logical patterns in the embedding space is still lacking.

Given a model $M$ of the query embedding method, we can distinguish two ways in which $M$ can exploit a logical pattern $P$. The first is the ability to \emph{predict} answers that are entailed by $P$. 
The second is the ability of model $M$ to \emph{explain} such predictions by inferring $P$ as an axiom that can be obtained from the geometrical relations between the objects in the embedding space.

Although query embedding methods have achieved great success in predicting query results over incomplete data, they fail to explain the learned knowledge because, to achieve better results, they endow geometric operations with neural network operations that impede explainability. For example, the method BetaE~\cite{BetaE} represents queries $q$ with multidimensional beta distributions $\embedding{q}$, and the set of answers to $q$ in $M$, denoted $\answers{q}{M}$, consists of the entities $e$ such that $\embedding{e} \in \embedding{q}$. A query $q(y) = \exists x (q_1(x) \land r(x,y))$, extending a given query $q_1(x)$ with a relation $r$, is embedded as a beta distribution $\embedding{q}$ that results from applying a neural network over the distribution $\embedding{q_1}$ for query $q_1$. The use of neural networks to represent relations hinders the inference of axioms like $\playsFor \circ \teamWon \sqsubseteq \athleteWon$. On the other hand, BoxEL~\cite{DBLP:conf/semweb/XiongPTNS22} and Box$^2$EL~\cite{DBLP:journals/corr/abs-2301-11118} embed description logic concepts, but are limited to reduced description logics ($\mathcal{EL}$ and $\mathcal{EL}^{++}$).


In this paper, we propose a novel query embedding model, {\OurMethod}, that can explain several logical patterns expressed as $\DLogic$ axioms learned for query answering tasks. Being more parameter-efficient than most existing approaches \cite{ConE,GQE,FuzzQE,querytobox}, {\OurMethod} reduces the dependency on neural networks by translating logical operators to a simpler algebraic structure. To achieve this explainability, {\OurMethod} embeds each $\DLogic$ concept as a multidimensional cone in the complex vector space, and relations as scaling and rotations of cones. Then, each logical operator is translated into geometric operations in embedding space. By doing so, our method can generate $\DLogic$ ontologies by learning query embeddings.

In summary, this paper makes the following contributions:
\begin{enumerate}
\item
  We formalize the notions of tree-form logical query in terms of $\DLogic$ concepts, and logical pattern in terms of $\DLogic$ axioms. We propose an algebra of cones in the complex plane, and theoretically identify the subset of the $\DLogic$ axioms that can be represented with the cone algebra models (Section~\ref{sec:ConceptsConeAlgebra}).
\item
  We present criteria where a multicone embedding expresses six different logical patterns, namely role containment, composition, transitivity, inverse, symmetry, and asymmetry (Section~\ref{sec:logical-patterns}).
\item
  We propose a novel method, {\OurMethod}, which embeds $\DLogic$ concepts as cones in the complex plane (Section~\ref{sec:AconE}). This technique allows us to leverage the rotation operator through Euler's formula while maintaining the geometric representation of concepts and concept operators to allow the explanation of logical patterns as $\DLogic$ axioms that the embedding explicitly encodes.
\item
  We show that modeling cones in the complex plane, coupled with the rotation operator as a complex product, is more parameter-efficient than modeling cones in the 2D real space with multi-layer neural operators (Section \ref{Analysis on Model Parameters}).
\item
  We create new datasets and dataset splits (Section 7.3) to conduct a more detailed analysis of the influence of patterns in complex query answering, providing a finer-grained evaluation, and more in-depth insights into the often-overlooked problem of pattern inference in complex query answering. 
\item
  Our experiments show that {\OurMethod} outperforms state-of-the-art baselines that represent query regions using vectors, geometries, or distributions, using geometric operations to model logical operations one-to-one (Section 7.2).
\end{enumerate}

\section{Related Work}

\paragraph{Logical patterns in knowledge graphs.}
KG embedding methods aim to learn KG representations that capture  latent structural and logical patterns \cite{TransE,DistMult,ComplEx,RotatE,PConE,TransH}.
In particular, RotatE \cite{RotatE} captures a broad range of logical patterns, such as symmetry, inversion, and composition, among others. 
While these KG embedding methods excel at predicting links, they cannot answer first-order logical queries.

BoxEL~\cite{DBLP:conf/semweb/XiongPTNS22,SEL} and Box$^2$EL~\cite{DBLP:journals/corr/abs-2301-11118} embed description logic concepts, but are limited to reduced description logics ($\mathcal{EL}$ and $\mathcal{EL}^{++}$). \cite{OLW-IJCAI20} proposes a method to embed $\mathcal{ALC}$ concepts using al-cones, which differs from our proposal in the geometrical representation of concepts. However, they do not propose a concrete geometry to embed relations (which is required for embedding tree-form queries).

\paragraph{Query answering.}
Path-based \cite{deeppath,LinSX18}, neural \cite{GQE,querytobox,ConE,BetaE,BiQE}, and neural-symbolic \cite{CQD,GNNQE,FuzzQE} methods have been developed to answer (subsets of) queries. Among these methods, geometric and probabilistic query embedding approaches \cite{GQE,querytobox,ConE,BetaE} provide an effective way to answer tree-form queries over incomplete and noisy KGs. This is done by representing entity sets as geometric objects or probability distributions, such as boxes \cite{querytobox}, cones \cite{ConE}, or Beta distribution \cite{BetaE}, and performing neural logical operations directly on them. The Graph Query Embedding (GQEs) \cite{GQE} was first proposed to answer only conjunctive queries via modeling the query $q$ as single vector $\embedding{q}$ through neural translational operators. However, modeling a query as a single vector limits the model's expressiveness in modeling multiple entities. Query2Box \cite{querytobox} remedies this flaw by modeling entities as points within boxes. This allows Query2Box to predict the intersection of entity sets as the intersection of boxes in vector space. ConE \cite{ConE} was proposed as the first geometry-based query embedding method that can handle negation via embedding the set of entities (query embedding) as cones in Euclidean space.

All of the above query embedding methods commonly apply multi-layer perceptron networks for selecting answer entities of atomic queries by relation and performing logical operations. Such methods suffer from two problems. Firstly, their ability to capture logical patterns in KGs remains unclear due to the limited explainability of neural networks. Secondly, a large number of parameters need to be trained for an outstanding model performance. Our proposed method, {\OurMethod}, overcomes these issues by providing a one-to-one mapping between logical and geometrical operators.


\section{Preliminaries}%
\label{preliminary}

\paragraph{The Description Logic $\DLogic$.}
We next present the standard $\DLogic$ syntax and semantics and assume standard semantics as defined in~\citet{DBLP:conf/dlog/2003handbook}. For these definitions, we assume three pairwise disjoint sets $\ConceptNames$, $\RoleNames$, and $\IndividualNames$, whose elements are respectively called \emph{concept names}, \emph{relation names}, and \emph{individual names}.

\begin{definition}[$\DLogic$ Concept Descriptions] 
  $\DLogic$ \emph{concept descriptions} $C$ and \emph{relation descriptions} $R$ are defined by the following grammar
  \begin{align*}
     C &\Coloneqq
      \top \mid A \mid \nominal{a} \mid \neg C \mid C \sqcap C
      \mid \exists R.C \\
     R &\Coloneqq r \mid r^-
  \end{align*}
  where the symbol $\top$ is a special concept name, and symbols $A$, $a$, and $r$ stand for concept names, individual names, and relation names, respectively. Concept descriptions $\nominal{a}$ are called \emph{nominals}.
  
  Given two concept descriptions $C$ and $D$, the expression $C \sqsubseteq D$ is a \emph{concept-axiom}. Given the relation descriptions $R,S,R_1, \dots, R_n$ (with $n \geq 1$), the expressions $\rho_1 \circ \dots \circ \rho_{n-1} \sqsubseteq \rho_n$, $\Disj{R}{S}$, $\Trans{R}$, $\Reflex{R}$, $\Irref{R}$, $\Symmetry{R}$, and $\Asymmetry{R}$ are \emph{relation-axioms}. Given two individual names $a,b \in \IndividualNames$, a concept description $C$ and a relation description $\rho$, $\ConceptAssertion{a}{C}$ is a \emph{concept-assertion} and $\RoleAssertion{\rho}{a}{b}$ is a \emph{relation-assertion}.
  
  We write $C \equiv D$ as an abbreviation for two axioms $C \sqsubseteq D$ and $D \sqsubseteq C$, and likewise for $\rho_1 \equiv \rho_2$. We write $\bot$, $C \sqcup D$, $\forall\rho.C$ as abbreviations for $\neg\top$, $\neg(\neg C \sqcap \neg D)$ and $\neg\exists\rho.\neg C$, respectively.
\end{definition}

\noindent An $\DLogic$ \emph{knowledge base} (or \emph{ontology}) $\KBase$ is a triple $(\RBox, \TBox, \ABox)$ where $\RBox$ is a finite set of relation-axioms called the \emph{RBox}, $\TBox$ is a finite set of concept-axioms called the \emph{TBox}, and $\ABox$ is a finite set of assertions called the \emph{ABox}.

An \emph{interpretation} $\Int$ is a pair $(\Delta^\Int, \cdot^\Int)$ consisting of a set $\Delta^\Int$, called the \emph{domain}, and a function $\cdot^\Int$ such that we have for each individual name $a \in \IndividualNames$, an element $a^\Int \in \Delta^\Int$; for each concept name $A \in \ConceptNames$, a subset $A^\Int \subseteq \Delta^\Int$; and for each relation name $r \in \RoleNames$, a relation $r^\Int \subseteq \Delta^\Int \times \Delta^\Int$. An interpretation $\Int$ is a \emph{model} of a knowledge base $\KBase$ if and only if all the axioms and assertions in $\KBase$ are satisfied according to the standard semantics defined in Baader et al.~\cite{DBLP:conf/dlog/2003handbook}. Given two knowledge bases $\KBase_1$ and $\KBase_2$, $\KBase_1$ \emph{entails} $\KBase_2$, denoted $\KBase_1 \models \KBase_2$, if and only if every model $\Int$ of $\KBase_1$ is also a model of $\KBase_2$.

\paragraph{Knowledge graphs and queries.} We next define knowledge graphs in terms of $\DLogic$ knowledge bases, and tree-form queries in terms of $\DLogic$ concepts.

\begin{definition}[Knowledge Graph]
  A knowledge graph $G$ is a $\DLogic$ knowledge base $\KBase$ whose RBox is empty, and its $\TBox$ contains a unique axiom $\top \sqsubseteq \{a_1\} \sqcup \cdots \sqcup \{a_n\}$, where $\{a_1,\dots,a_n\}$ is the set of all individuals names occurring in the ABox. This axiom is called \emph{domain-closure assumption}.
\end{definition}

\begin{definition}[Tree-form query]
  Given a knowledge graph $G$, and an individual name $x$ that does not occur in $G$, a
  \emph{tree-form query} $q$ is an assertion $\ConceptAssertion{C}{x}$ where $C$ is a $\DLogic$ concept description. The answers to query $q$, are all individuals $a$ occurring in $G$ such that $G \models \ConceptAssertion{C}{a}$.
\end{definition}

Informally, the query answering task over incomplete data consists of predicting answers to a query $q$ over a knowledge graph pattern $G$ given only a subset $G' \subset G$. Knowledge graph embeddings to generalize the knowledge on $G'$ to predict answers on $G$.


\section{$\DLogic$ concepts and Cone Algebra}%
\label{sec:ConceptsConeAlgebra}

In this section, we present an algebra of cones in the complex plane, called the {\ConeAlgebra}, we show the correspondence between this algebra and $\DLogic$ concepts, and we identify the subset of the $\DLogic$ axioms that are expressible with this algebra.

\begin{definition}[Cone]
  A \emph{cone} $\Cone(\alpha, \beta)$ is a region in the complex plane $\Complex$ determined by a pair of angles $(\alpha,\beta) \in \Real^2$ as follows:
  \[
    \Cone(\alpha,\beta) = \{ e^{i\theta} \st \alpha \leq \theta \text{ and } \theta \leq \beta \}.
  \]
  The \emph{empty cone}, denoted $\EmptyCone$, is the cone such that $\alpha > \beta$. A \emph{singleton cone} with angle $\alpha$, denoted $\SingletonCone{\alpha}$, is a cone such that $\alpha = \beta$. A \emph{proper cone}, denoted $\ProperCone{\alpha}{\beta}$, is a cone where $\alpha + 2\pi > \beta$. A \emph{full cone}, denoted $\FullCone$, is the cone such that $\alpha + 2\pi \leq \beta$.
\end{definition}

Notice that $e^{i\theta} = e^{i\theta + 2k\pi}$, for every natural number $k$. Thus, the same cone can be determined with multiple combinations of angles.

Notice that the intersection is not closed on the set of cones. Indeed, the region $\Cone(0, \frac{3}{2}\pi) \cap \Cone(\pi, \frac{5}{2}\pi)$ is not a cone. This hinders the use of cones to represent $\DLogic$ concepts. To overcome this limitation, we can embed concepts on sets of cones.

\begin{definition}[Multicone algebra]
  A \emph{multicone} $\MCone(\SetCones)$ is a region determined by a set of cones $\SetCones$ as follows:
  \[
    \MCone(\SetCones) = \bigcup_{\Cone(\alpha,\beta) \in \SetCones} \Cone(\alpha, \beta)
  \]
  The \emph{multicone algebra} is the algebra over the set of multicones defined by the binary operations $\cap$ and $\cup$ that are defined as the set operations over the multicon regions.
\end{definition}

\begin{proposition}
    The operations $\cup$ and $\cap$, are commutative, associative, and mutually distributive, and their identity elements are $\EmptyCone$ and $\FullCone$, respectively.
\end{proposition}


\begin{definition}[Rotation algebra]%
  \label{def:rotation-algebra}
  Given a triple $(\theta, \gamma, \delta) \in \Real^3$ where $0 \leq \gamma \leq 2\pi$, a \emph{rotation} $\Rotation{\theta}{\gamma}{\delta}$ is a function that maps every singleton and proper cone $\Cone(\alpha, \beta)$ to the cone $\Cone(\alpha', \beta')$ such that
  \begin{align*}
    \beta' + \alpha' &= \theta(\beta + \alpha),&
    \beta' - \alpha' &= \gamma(\beta - \alpha) + \delta,
  \end{align*}
  and that maps the empty and the full cone to themselves:
  \begin{align*}
    \Rotation{\theta}{\gamma}{\delta}(\EmptyCone) &= \EmptyCone,&
    \Rotation{\theta}{\gamma}{\delta}(\FullCone) &= \FullCone.
  \end{align*}
  The rotation parameters $\theta$, $\gamma$, and $\delta$ are called the \emph{rotation angle} and the \emph{aperture factor}, and the \emph{aperture adding}. We call \emph{aperture-multiplicative} and \emph{aperture-additive} rotations to rotations of the respective forms $\Rotation{\theta}{\gamma}{0}$ and $\Rotation{\theta}{1}{\delta}$.
  
  We call \emph{rotation algebra} to be the algebraic structure whose ground set is the set of rotations over cones, and has a binary operation $\circ$ denoting function composition (i.e., $(f \circ g)(x) = g(f(x))$).

  We write $\Rotation{\theta}{\gamma}{\delta}^- = \Rotation{-\theta}{\frac{1}{\gamma}}{-\delta}$, and
  with a slight abuse of notation, given a multicone $\MCone(\SetCones)$, we write
  \[
    \Rotation{\theta}{\gamma}{\delta}(\MCone(\SetCones)) =\MCone(\{ \Rotation{\theta}{\gamma}{\delta}(\Cone(\alpha, \beta)) \st \Cone(\alpha, \beta) \in \SetCones \}).
  \]
\end{definition}


\begin{definition}[Multicone Embedding]%
  \label{def:multicone-embedding}
  A \emph{multicone embedding} is a function $\Embedding$ that maps each individual name $a \in \IndividualNames$ to a complex number $a^{\Embedding} = e^{i\theta} \in \FullCone$, each concept name $A \in \ConceptNames$ to a multicone $A^{\Embedding}$, and each relation name $r \in \RoleNames$ to a rotation $r^{\Embedding} = \Rotation{\theta}{\gamma}{\delta}$. The \emph{multicone embedding of a knowledge graph} $G$ is a multicone embedding restricted to the individual names and relation names occurring in knowledge graph $G$.
\end{definition}

\begin{table}[t]
  \caption{Multicone Embedding Semantics.}%
  \label{table:embedding-concept-descriptions}
  \centering
  \begin{adjustbox}{max width=\hsize}
  \begin{tabular}{ll}
    \toprule
    Concept
    & Semantics \\ \midrule
    $\top$
    & $\FullMCone$ \\
    $\nominal{a}$
    & $\MCone(\{\SingletonCone{\theta}\})$ where $a^{\Embedding} = e^{i\theta}$ \\
    $C \sqcap D$
    & $C^{\Embedding} \cap D^{\Embedding}$ \\
    $\neg C$
    & $\FullMCone \setminus C^{\Embedding}$ \\
    $\exists R.C$
    & $R^{\Embedding}(C^{\Embedding})$ \\
    $r^-$
    & $(r^{\Embedding})^-$ \\
    $R \circ S$
    & $R^{\Embedding} \circ S^{\Embedding}$ \\
    \midrule
    $\ConceptAssertion{A}{a}$
    & $a^{\Embedding} \in A^{\Embedding}$ \\
    $\RoleAssertion{R}{a}{b}$
    & $(a^\Int, b^{\Embedding}) \in R^{\Embedding}$ \\
    \midrule
    $C \sqsubseteq D$
    & $C^{\Embedding} \subseteq D^{\Embedding}$ \\
    $R_1 \circ \cdots \circ R_n\sqsubseteq S$
    & $(R_1 \circ \dots \circ R_n)^{\Embedding} \subseteq S^{\Embedding}$ \\
    $\Disj{R}{S}$
    & $R^{\Embedding} \cap S^{\Embedding} = \emptyset$ \\
    $\Trans{R}$
    & $(R \circ R)^{\Embedding} \subseteq R^{\Embedding}$ \\
    $\Reflex{R}$
    & $\{(a,a) \mid a \in \IndividualNames \} \subseteq R^{\Embedding}$ \\
    $\Irref{R}$
    & $\{(a,a) \mid a \in \IndividualNames \} \cap R^{\Embedding} = \emptyset$ \\
    $\Symmetry{R}$
    & for all $a,b \in \IndividualNames$, if $(a,b) \in R^{\Embedding}$ then $(b,a) \in R^{\Embedding}$ \\
    $\Asymmetry{R}$
    & for all $a,b \in \IndividualNames$, if $(a,b) \in R^{\Embedding}$ then $(b,a) \notin R^{\Embedding}$ \\
    \bottomrule
  \end{tabular}
  \end{adjustbox}
\end{table}

Given a multicone embedding $\Embedding$, Table~\ref{table:embedding-concept-descriptions} defines the multicone $C^{\Embedding}$ corresponding to a $\DLogic$ concept description $C$. The definition of the multicone semantics for relations, assertions and axioms is straightforward. For example, the embedding of $R \circ S$ is the relation $\{ (a,c) \st (a,b) \in R^{\Embedding},\, (b,c) \in S^{\Embedding} \}$, and an $\Embedding \models \ConceptAssertion{C}{a}$ if and only if $a^{\Embedding} \in C^{\Embedding}$,
$\Embedding \models \RoleAssertion{R}{a}{b}$ if and only if $(a^{\Embedding}, b^{\Embedding}) \in R^{\Embedding}$, and $C \models D$ if and only if $C^{\Embedding} \subseteq D^{\Embedding}$.

\paragraph{The full cone.}
Since $\FullCone$ is an absorbing element for the rotation algebra, multicone embeddings infer wrong axioms $\top \sqsubseteq \exists R.\top$.

\paragraph{The rotation commutativity.}
In certain cases, the order in which rotations are applied does not affect the resulting cone.

\begin{proposition}
  Given two aperture-multiplicative rotations $\Rotation{\theta_1}{\gamma_1}{0}$ and $\Rotation{\theta_2}{\gamma_2}{0}$ such that $\gamma_1 \leq 1$ and $\gamma_2 \leq 1$, then
  \[
    \Rotation{\theta_1}{\gamma_1}{0} \circ \Rotation{\theta_2}{\gamma_2}{0} =
    \Rotation{\theta_2}{\gamma_2}{0} \circ \Rotation{\theta_1}{\gamma_1}{0}.
  \]
\end{proposition}


The issue with the commutativity is that if two relation names $r$ and $s$ are embedded with aperture-multiplicative rotations $\Rotation{\theta_r}{\gamma_r}{0}$ and $\Rotation{\theta_s}{\gamma_s}{0}$ with $\gamma_r \leq 1$ and $\gamma_s \leq 1$, then we can infer that $r \circ s \equiv s \circ r$. So, the commutativity of these rotations may introduce a bias into the models that can lead to wrong predictions.

This bias favoring commutativity does not hold if $\gamma > 1$, as the following counter example shows.
\begin{align*}
  &\textstyle (\Rotation{0}{2}{0} \circ \Rotation{0}{\frac{1}{2}}{0})(\Cone(0, \pi)) =
  \FullCone,\\
  &\textstyle (\Rotation{0}{\frac{1}{2}}{0} \circ \Rotation{0}{2}{0})(\Cone(0, \pi)) =
  \Cone(0, \pi).    
\end{align*}
However, rotations with $\gamma > 1$ generate larger cones (sometimes full cones), and thus can lead to a reduced prediction accuracy.

Aperture-additive rotations are not commutativity because, for every $\delta \neq 0$, we can find a cone $\Cone$ such that $\Rotation{\theta}{1}{\delta}(\Cone)$ is either $\FullCone$ or $\EmptyCone$ (which are absorving elements). However, under certain conditions, commutativity holds. For example, if $u$, $v$, and $w$ are three positive real numbers with $u + v \leq w$, and $\Rot_1 = \Rotation{\theta_1}{1}{u}$, $\Rot_2 = \Rotation{\theta_2}{1}{v}$, $\Rot_3 = \Rotation{\theta_1}{1}{-u}$, $\Rot_4 = \Rotation{\theta_2}{1}{-v}$, then
\begin{align*}
  (\Rot_1 \circ \Rot_2)(\Cone(0, 2\pi - w)) &= (\Rot_2 \circ \Rot_1)(\Cone(0, 2\pi - w)), \\
  (\Rot_3 \circ \Rot_4)(\Cone(0, w)) &= (\Rot_4 \circ \Rot_3)(\Cone(0, w)).
\end{align*}
Under these circumstances, there is a bias that favors the inference of axioms $\exists R_1(\exists R_2.C) \equiv \exists R_2(\exists R_1.C)$. Hence, both aperture-multiplicative and aperture-additive rotations are biased.

\paragraph{Distributivity of the existential over the disjunction.}
To guarantee that the operations are closed, we needed to extend the embeddings to multicones. However, multicones do not satisfy a basic property. The full multicone $\FullMCone$ can be understood as two possible multicones, namely $\MCone(\{\FullCone\})$ and $\MCone(\{\Cone(0,\pi), \Cone(\pi, 2pi)\})$. However, the application of the rotation $R^{\Embedding} = \Rotation{\theta}{\frac{1}{2}}{0}$ returns different multicones for both representations of the multicone. Since a multicone $\MCone(\{\Cone_1,\Cone_2\})$ is equal to $\Cone_1 \cup \Cone_2$, the tautology $\exists R.(C \sqcup D) \equiv \exists R.C \sqcup \exists R.D$ does not hold in the embedding space. Indeed, it only holds for positive aperture-additive rotations.

\begin{proposition}\label{prop:distrib}
  The axiom $\exists R.(C \sqcup D) \equiv \exists R.C \sqcup \exists R.D$ holds in multicone embeddings for rotations $\Rotation{\theta}{\gamma}{\delta}$ if $\gamma = 1$ and $\delta \geq 0$.
\end{proposition}


\section{Expressing Logical Patterns}%
\label{sec:logical-patterns}

In this section, we present how logical patterns are expressed in the multicone embedding space. Given the relation names $r_1, r_2, r_3 \in \RoleNames$, correspond to the following $\DLogic$ axioms:
\begin{align*}
  &r_1 \sqsubseteq r_2 &&\text{(role containment)},&
  &r_1 \circ r_2 \sqsubseteq r_3 &&\text{(composition)},\\[-3pt]
  &\Trans{r_1} &&\text{(transitivity)},&
  &r_1 \equiv \inverse{r_2} &&\text{(inverse)},\\[-3pt]
  &\Symmetry{r_1} &&\text{(symmetry)},&
  &\Asymmetry{r_1} &&\text{(asymmetry)}.
\end{align*}

\noindent\paragraph{Role containment pattern.}
Given two rotations $R^{\Embedding} = \Rotation{\theta_1}{\gamma_1}{0}$ and $S^{\Embedding} = \Rotation{\theta_2}{\gamma_2}{0}$ with $\theta_1 \neq \theta_2$ and $\gamma_1 \neq \gamma_2$, it never hold that $R^{\Embedding} \sqsubseteq S^{\Embedding}$. Indeed, we can always consider a cone with a sufficiently small aperture angle (e.g., a singleton cone) to show that this inclusion does not hold. However, for rotations $R^{\Embedding} = \Rotation{\theta_1}{1}{\delta_1}$ and $S^{\Embedding} = \Rotation{\theta_2}{1}{\delta_2}$ the inclusion can hold, even with distinct values for $\theta_1$ and $\theta_2$.

\begin{proposition}\label{prop:role-containment-additive}
  Given two rotations $R^{\Embedding} = \Rotation{\theta_1}{1}{\delta_1}$ and $S^{\Embedding} = \Rotation{\theta_2}{1}{\delta_2}$, the axiom $R \sqsubseteq S$ holds if $\theta_2 + \frac{\delta_2}{2} \geq \theta_1 + \frac{\delta_1}{2}$ and $\theta_2 - \frac{\delta_2}{2} \leq \theta_1 - \frac{\delta_1}{2}$
\end{proposition}

\begin{proposition}\label{prop:role-containment-product}
  Given two rotations $R^{\Embedding} = \Rotation{\theta}{\gamma_1}{\delta_1}$ and $S^{\Embedding} = \Rotation{\theta}{\gamma_2}{\delta_2}$, the axiom $R \sqsubseteq S$ holds if $\gamma_1 \leq \gamma_2$ and $\delta_1 \leq \delta_2$.
\end{proposition}

\paragraph{Composition and transitivity patterns.}
The rules for composition and transitivity are a corollary of Proposition~\ref{prop:role-containment-additive} and Proposition~\ref{prop:role-containment-product}. For space limitations, we omit the conditions for the Transitivity pattern. The conditions can be obtained from the equivalence of axioms $\Trans{R}$ and $R \circ R \sqsubseteq R$.

\begin{corollary}
  Given three rotations $R_1^{\Embedding} = \Rotation{\theta_1}{1}{\delta_1}$, $R_2^{\Embedding} = \Rotation{\theta_2}{1}{\delta_2}$, and $R_3^{\Embedding} = \Rotation{\theta_3}{1}{\delta_3}$, the axiom $R_1 \circ R_2 \sqsubseteq R_3$ holds if $|\theta_2 - (\theta_1 + \theta_2)| \leq \delta_3 - (\delta_1 + \delta_2)$, $\delta_1 \geq 0$, $\delta_2 \geq 0$, and $\delta_3 \geq 0$.
\end{corollary}

\begin{corollary}
  Given $R_1^{\Embedding} = \Rotation{\theta}{\gamma_1}{\delta_1}$, $R_2^{\Embedding} = \Rotation{\theta}{\gamma_2}{\delta_2}$, and $R_3^{\Embedding} = \Rotation{\theta}{\gamma_3}{\delta_3}$, the axiom $R_1 \circ R_2 \sqsubseteq R_3$ holds if $\gamma_1 \gamma_2 \leq \gamma_3$, $\delta_1 + \delta_2 \leq \delta_3$, $\delta_1 \geq 0$, and $\delta_2 \geq 0$.
\end{corollary}

\paragraph{Inverse pattern.}
Because inverses cannot generally be found, we cannot define inverse patterns for relations. However, inverse patterns still can be inferred under the circumstances where the commutativity is held, which is discussed in Section~\ref{sec:ConceptsConeAlgebra}.


\paragraph{Symmetry and asymmetry patterns.}
The symmetry pattern remains consistent when a single cone is rotated twice, resulting in the same cone. This happens when two rotations complete the circle. Similarly, the asymmetry pattern holds when two rotations do not coincide with a circle.

\begin{proposition}
  Given a rotation $R^{\Embedding} = \Rotation{\theta}{\gamma}{\delta}$ with $\gamma \geq 1$ and $\rho \geq 0$, if there is a natural number $k$ such that $2\theta + \frac{\gamma\rho + \rho}{2} \geq 2k\pi$ and $2\theta - \frac{\gamma\rho + \rho}{2} \leq 2k\pi$, then the axiom $\Symmetry{r}$ holds. 
\end{proposition}

\section{Tree-form Query Answering with {\OurMethod}}%
\label{sec:AconE}

To accommodate the learning of logical patterns in query answering over KGs, we propose a new model, {\OurMethod}, which embeds KGs according to the multicone embedding (Definition~\ref{def:multicone-embedding}).
Our approach distinguishes itself from ConE \cite{ConE}, which represents cone embeddings using two-dimensional vectors in real space. We reframe these embeddings in the complex plane. By doing so, we introduce an inductive bias that facilitates the learning of logical patterns. This bias is achieved by combining the embeddings with relational rotations on cones, enhancing the model's ability to capture complex relationships and patterns. In {\OurMethod}, the embedding of a tree-form query $q = \ConceptAssertion{C}{x}$ is parameterized by a pair $\embedding{q} = (\UperBound, \LowerBound) \in \Complex^d$, where $d$ is the embedding dimension, $|\UperBound|_2 = \VecOne$, $|\LowerBound|_2 = \VecOne$ with $|\cdot|_2$ being the L2 norm. The vectors $\UperBound$ and $\LowerBound$ represent the counter-clockwise upper and lower boundaries of the cone, such that 
\begin{align}
  \UperBound &= e^{i\UperAngle} = e^{i\left(\AxAngle + \frac{\ApAngle}{2}\right)}, &
  \LowerBound &= e^{i\UperAngle} = e^{i\left(\AxAngle - \frac{\ApAngle}{2}\right)},
\end{align}
where $\AxAngle \in [-\pi,\pi)^d$ represents the angle of the symmetry axis of the cone and $\ApAngle \in [0,2\pi]^d$ represents the cone aperture.

For each index $j$ with $1 \leq j \leq d$, we write $\VectorComponent{\embedding{v}}{j}$ for the value of the $j$-th component of a vector $\embedding{v}$. The i-th component of $\embedding{q}$ represents the proper cone $\Cone(\LowerAngleJ{j}, \UperAngleJ{j})$ if $\ApAngleJ{j} > 0$ and $\ApAngleJ{j} < 2\pi$, the full cone $\FullCone$ if $\ApAngleJ{j} = 2\pi$, and the singleton cone $\SingletonCone{\LowerAngleJ{j}}$ if $\ApAngle = 0$. Notice that we do not provide an encoding for the empty cone because interesting queries are satisfiable. The queries are thus modeled as multidimensional cones according to Definition~\ref{def:multicone-embedding}, and each entity answering the query is modeled as a vector $\EntityVector = e^{i\EntityAngle}$ in the cone parametrized with the query embedding.

\begin{figure}[t]
    \vspace{-10pt}
    \centering
    \includegraphics[width=\linewidth]{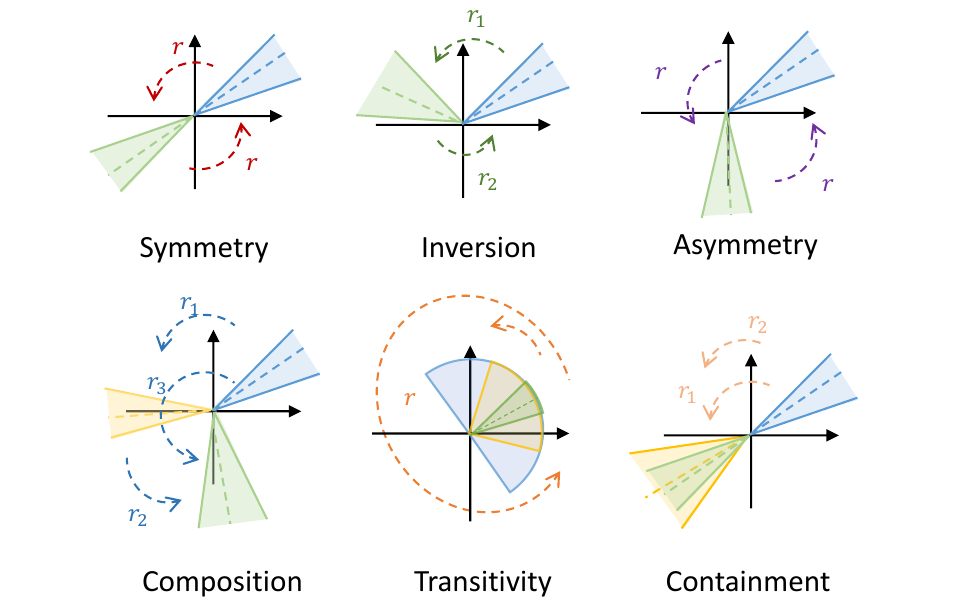}
    \caption{Logical patterns captured by AConE in a single dimension}
    \label{fig:logical_patterns in AConE}
    \vspace{15pt}
\end{figure}
\subsection{Geometric Operators}
To answer tree-form queries, {\OurMethod} translates the concept constructors $\exists$, $\sqcap$, $\sqcup$, and $\neg$ into corresponding geometric operators, i.e. \emph{relational rotation} $\GeoOpExists$, \emph{intersection} $\GeoOpConj$, \emph{union} $\GeoOpDisj$, and \emph{negation} $\GeoOpNeg$, in the complex plane. {\OurMethod} derives the final query embedding by executing these geometric operators along the computation graph of the query. We next describe these geometric operators.

\paragraph{Relational Rotation.}
Given a set of entities $\mathcal{S} \subset \IndividualNames$ and a relation $r \in \RoleNames$, the transformation operator for $r$ selects the entities $\mathcal{S}' = \{e' \in \IndividualNames \st G \models r(e, e'),\, e \in \mathcal{S}\}$. To this end, we model $r$ with a vector $\rotation = (\rotationU, \rotationL) \in \Complex^{2\times d}$ encoding a counterclockwise rotation on query embeddings about the complex plane origin such that $|\rotationU| = \VecOne$ and $|\rotationL| = \VecOne$. The rotation $\rotation$ transforms a query embedding $\embedding{q}  = (\UperBound, \LowerBound)$ into a query embedding $\GeoOpExists(\embedding{q}, \rotation) = \embedding{q}' = (\UperBound', \LowerBound')$ where 
\begin{align}
  \UperBound' &= \UperBound \circ \rotationU,&
  \LowerBound' &= \LowerBound \circ \rotationL,
\end{align}
where $\circ$ is the Hadamard (element-wise) product (i.e., for each component $j$, $\UperBoundJ{j}' = \UperBoundJ{j} \cdot \rotationUJ{i}$ and
$\LowerBoundJ{j}' = \LowerBoundJ{j} \cdot \rotationLJ{i}$). If a pair $(\rotationUJ{j}, \rotationLJ{j})$ is seen as a cone, with axis angle $\AxRAngleJ{j}$ and aperture angle $\ApRAngleJ{j}$, then the rotation corresponds to the aperture-additive rotation $\Rotation{\AxRAngleJ{j}}{1}{\ApRAngleJ{j}}$ (see Definition~\ref{def:rotation-algebra}). That is, the axis and aperture angles of $\embedding{q}'$ are
$\AxAngle' = \AxAngle + \AxRAngle$, and $\ApAngle' = \ApAngle + \ApRAngle$. Figure \ref{fig:logical_patterns in AConE} illustrates how AConE models the logical patterns described in Section \ref{sec:logical-patterns}, using the example on a single dimension.

%
\paragraph{Intersection.}
The intersection of a set of query embeddings \(Q=\{\embedding{q_1},...,\embedding{q_n}\}\), denoted $\GeoOpConj(Q) = \embedding{q}'$ is defined with the permutation-invariant functions $\SemanticAverage$ and $\CardMin$~\cite{ConE}, which calculate the center $\AxAngle'$ and the aperture $\ApAngle'$ of the resulting cone encoded by $\embedding{q}'$ as follows:
\begin{equation}
\begin{split}
    \AxAngle' &= \SemanticAverage(\{\embedding{q_k}\}_{k=1}^{n}),\\
    \ApAngle' &= \CardMin(\{\embedding{q_k}\}_{j=1}^{n}).
\end{split}
\end{equation}
$\SemanticAverage$ is expected to approximate the axis of the cone resulting from the intersection of the input cones. $\CardMin$ predicts the aperture $\ApAngle'$ of the intersection set such that $[\ApAngle']_{j}$ should be no larger than the aperture of any cone $\embedding{q_i} \in Q$, since the intersection set is the subset of all input entity sets. Both functions, $\SemanticAverage$ and $\CardMin$, use a neural network to improve the results. We extend the details of them in Appendix D in the supplementary material \cite{he2024generatingsroiontologiesknowledge}.

\paragraph{Union.}
Given a set of query embeddings \(Q = \{\embedding{q_1}, \dots, \embedding{q_n}\}\), the union operator $\GeoOpDisj(Q)$ returns the set $Q$. Intuitively, for each $k$ where $1 \leq k \leq d$, the set $Q$ encodes a multicone embedding for the query $(q_1 \sqcup \dots \sqcup q_n)(x)$, which represents the multicone $\MCone(\Cone_1,\dots,\Cone_n)$ where $\Cone_i$ is the cone encoded in $\VectorComponent{\embedding{q_i}}{k}$.

Notice that multicones cannot be further used as input of other operations because all {\OurMethod} operations are applied on cones. To compute queries, we thus follow \cite{querytobox}, translating queries into disjunctive normal form, so we only perform the disjunction operator in the last step in the computation graph.

\paragraph{Negation.}
The negation of a cone $\embedding{q} = (\UperBound, \LowerBound)$, denoted $\GeoOpNeg(\embedding{q})$, is the cone $\embedding{q}'$ that contains the entities in the complement of $\embedding{q}$, that is, $\embedding{q}' = (\LowerBound, \UperBound)$.

\subsection{Optimization}

\paragraph{Learning Objective.}
Given a set of training samples, our goal is to minimize the distance between the query embedding $\QueryEmbedding$ and the answer entity vector $\PositiveSample$, while maximizing the distance between $\QueryEmbedding$ and its negative samples. Thus, we define our training objective as 
\begin{equation}
L = -\log \sigma (\gamma - \CombDistance(\PositiveSample, \QueryEmbedding)) -
    \frac{1}{k}\sum_{i=1}^{k} \log\sigma(d(\NegativeSample{i},\QueryEmbedding) - \gamma),
\end{equation}
where $\CombDistance$ is the \emph{combined distance} defined below, $\gamma$ is a margin, $\PositiveSample$ is a positive entity, and $\NegativeSample{i}$ is the $i$-th negative entity, $k$ is the number of negative samples, and $\sigma$ represents the sigmoid function.

\paragraph{\textbf{Combined Distance.}} Inspired by \cite{ConE,querytobox}, the distance between $\QueryEmbedding$ and $\EntityVector$ is defined as a combination of inside and outside distances, $d_{com}(\QueryEmbedding, \EntityVector) = \OutDistance(\QueryEmbedding, \EntityVector) + \lambda \InDistance(\QueryEmbedding, \EntityVector)$:
\begin{align}
    \OutDistance(\QueryEmbedding, \EntityVector)
    &= \min\{\|\UperBound - \EntityVector\|_{1}, \| \LowerBound - \EntityVector\|_{1}\},\\
    \InDistance(\QueryEmbedding, \EntityVector)
    &= \min\{\|\ConeAxis - \EntityVector\|_{1}, \|\UperBound - \ConeAxis\|_{1}\},
\end{align}
where $\|\cdot\|_{1}$ is the $L_1$ norm, $\ConeAxis$ represents the cone center, and $\lambda \in (0,1)$. Note that $\CombDistance$ can only be used for measuring the distance between a single query embedding and an answer entity vector. Since we represent the disjunctive queries in Disjunctive Normal Form as a set of query embeddings $Q$, the distance between the answer vector and such set of embeddings is the minimum distance:
\begin{equation}
    \CombDistance(Q, \EntityVector) = \min\{\CombDistance(\QueryEmbedding, \EntityVector) \st \QueryEmbedding \in Q \}.
\end{equation}



\section{Experiments and Analysis}

%
In this section, we answer the following research questions with experimental statistics and corresponding case analyses. 
\textbf{RQ1}: how well does AConE improve query answering on incomplete KGs? \textbf{RQ2}: How is the improvement of results related to the capturing of logical patterns?

\subsection{Experimental Setup}
\begin{figure}[t]
  \centering
  \includegraphics[width=\hsize]{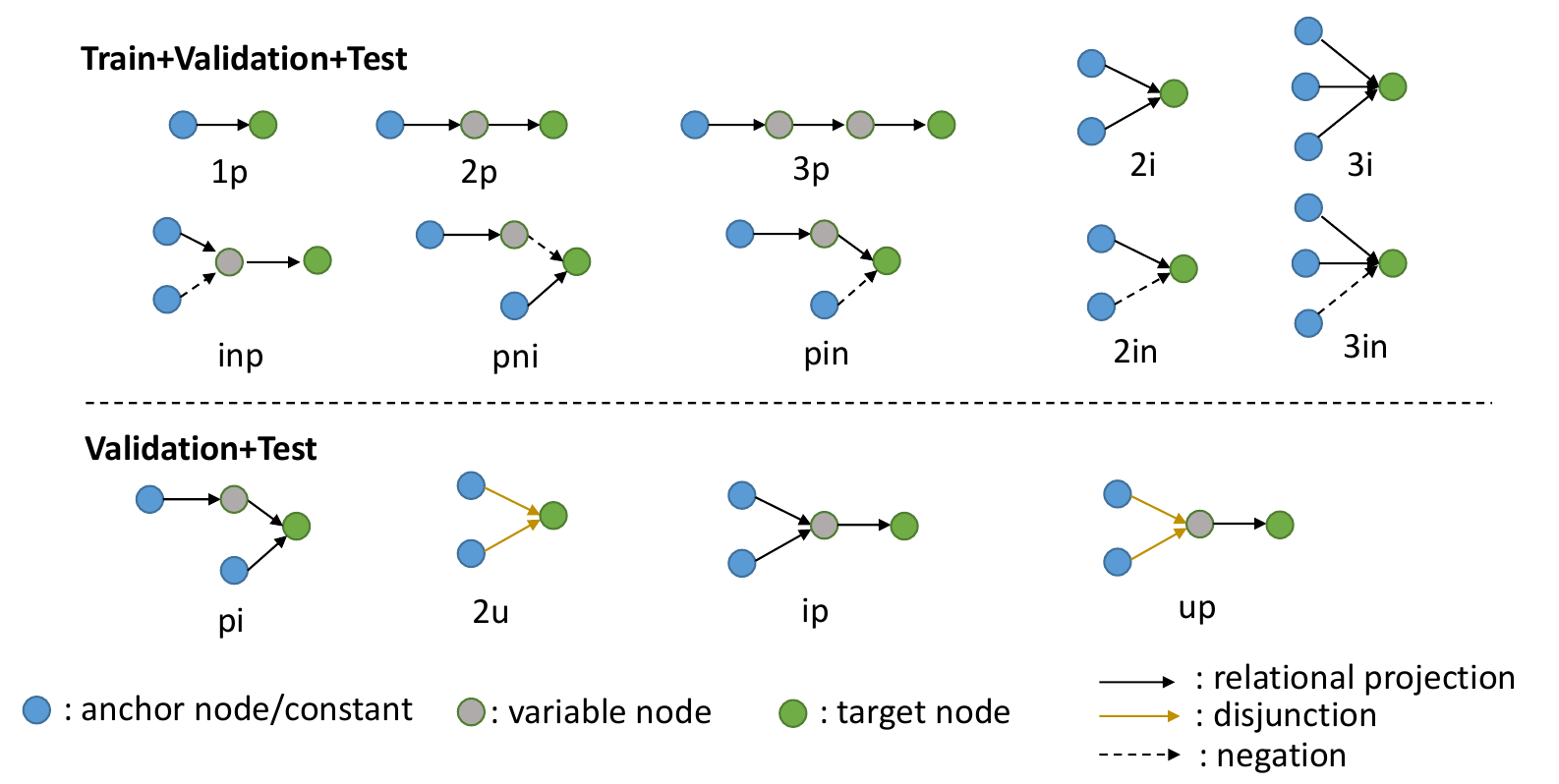}
  \caption{Fourteen types of queries used in the experiments. $\mathsf{p}$ represents an edge to another entity labeled with a relation name $r$ (i.e., an operation $\exists r$), $\mathsf{i}$ represents the intersection ($\sqcap$), $\mathsf{u}$ represents the disjunction ($\sqcup$), and $\mathsf{n}$ represents the negation ($\neg$).}
  \label{query structures}
  \vspace{15pt}
\end{figure}

\paragraph{\textbf{Dataset and Query Structure.}} For a fair comparison with the baseline models, we use the same query structures and logical query datasets from NELL-QA \cite{deeppath} and WN18RR-QA \cite{LinE}, and the open-sourced framework created by \cite{BetaE} for logical query answering tasks. Figure \ref{query structures} illustrates all query structures in experiments. We trained our model using 10 specific query structures and tested it on all 14 query structures to evaluate the generalization ability of our model regarding unseen query structures. More experimental details are in the supplementary material \cite{he2024generatingsroiontologiesknowledge}. Our code is available at \cite{code}.

\paragraph{\textbf{Evaluation Metrics.}}
We use Mean Reciprocal Rank (MRR) as the evaluation metric. Given a sample of queries Q, MRR represents the average of the reciprocal ranks of results, $\operatorname{MRR} = \frac{1}{|Q|}\sum_{i=1}^{|Q|}\frac{1}{\operatorname{rank}_{i}}$.

\subsection{RQ1: how well does AConE improve query answering on incomplete KGs?}
We evaluate AConE on two logical query-answering benchmark datasets that include a variety of complex logical patterns: NELL-QA created by \cite{BetaE} and WN18RR-QA by \cite{LinE}. AConE is compared with various query embedding models that also define query regions as vectors, geometries, or distributions, including GQE \cite{GQE}, Query2Box (Q2B) \cite{querytobox}, BetaE \cite{BetaE}, ConE \cite{ConE}, and LinE \cite{LinE}. To ensure a fair comparison, we indeed referenced the results of all baseline models from corresponding papers. In the case of ConE's performance on the WN18RR dataset, we conducted our own experiments due to the absence of previously reported results. This involved rerunning ConE with an optimized search for hyperparameters to generate the relevant data.

\paragraph{\textbf{Main Results.}} Table \ref{EPFOL} summarizes the performance of all methods on answering various query types. Compared with baselines that can only model queries without negation. Overall, AConE (aperture-additive) outperforms baseline methods on the majority of query types in Figure \ref{query structures} while achieving competitive results on the others. Specifically, AConE consistently achieves improvements on non-negation queries on all datasets. Furthermore, our model brings notable improvement over baseline models on the WN18RR-QA dataset, where the average accuracy of AConE is 18.35$\%$ higher than that of the baseline models.

\begin{table*}[t]
  \caption{MRR results (\%) of AConE, LinE (the results of LinE are only available on WN18RR-QA), ConE, BETAE, Q2B, and GQE on answering tree-form ($\exists, \land, \lor, \neg$) queries on datasets NELL-QA, and WN18RR-QA. The best statistic is highlighted in bold, while the second best is highlighted in underline.}
  \label{EPFOL}
  \centering
  \begin{tabular}{llllllllllllllll}
    \toprule
    \textbf{Dataset} & \textbf{Model} & \textbf{1p} & \textbf{2p} & \textbf{3p} & \textbf{2i} & \textbf{3i} & \textbf{pi} & \textbf{ip} & \textbf{2u} & \textbf{up} & \textbf{2in} & \textbf{3in} & \textbf{inp} & \textbf{pin} & \textbf{pni}\\
    \midrule
                     & GQE & 18.0 & 4.5 &2.7 &19.3& 23.9 & 9.9 &10.6 &2.3 &3.5 &-& -& -& -& -\\
                     & Q2B& 22.4 &4.6 &2.3& 25.6& 41.2 &13.2 &11.0 &2.9 &3.4& - &- &-& - &-\\
    WN18RR-QA & BetaE &44.1& 9.8 &3.8 &57.2& 76.2 &32.6 &17.9 & 7.5 &5.3 &12.7& 59.9& 5.1& 4.0& 7.4\\
                     & LinE & 45.1& 12.3 &6.7 &47.1 &67.1& 24.8& 14.7& 8.4& 6.9 &12.5 &60.8& 7.3 &5.2& \underline{7.7}\\
                     & ConE & \underline{46.8} & \underline{14.5} & \underline{9.3} & \underline{59.0} & \underline{83.9} & \underline{33.6} &\underline{18.7} & \underline{10.0} & \underline{9.8} & \underline{13.9} & \underline{61.8} & \underline{10.6} & \underline{7.3} & 7.6 \\
                     & AConE &  \textbf{50.9} & \textbf{17.6} & \textbf{9.9} & \textbf{70.5} & \textbf{89.0} &\textbf{38.9} & \textbf{29.6 }& \textbf{18.4} & \textbf{14.0} & \textbf{18.3} & \textbf{70.1} & \textbf{13.4} & \textbf{7.6}& \textbf{10.4} \\
    \midrule
                     & GQE & 33.1 & 12.1 & 9.9 & 27.3 & 35.1 & 18.5 & 14.5 & 8.5 & 9.0 & - & - & - & - & -\\
                     & Q2B & 42.7 & 14.5 & 11.7 & 34.7 & 45.8 & 23.2 & 17.4 & 12.0 & 10.7 & - & - & - & - & -\\
    NELL-QA & BetaE & 53.0 & 13.0 & 11.4 & 37.6 & 47.5 & 24.1 & 14.3 & 12.2 & 8.5 & 5.1 & \underline{7.8} & \underline{10.0} & 3.1 & 3.5\\
                     & ConE & \underline{53.1} & \underline{16.1} & \underline{13.9} & \underline{40.0} & \underline{50.8} & \textbf{26.3} & \underline{17.5} & \underline{15.3} & \underline{11.3} & \textbf{5.7} & \textbf{8.1} & \textbf{10.8} & \textbf{3.5} & \textbf{3.9}\\
                     & AConE & \textbf{54.5} & \textbf{17.7} & \textbf{14.4} & \textbf{41.9} & \textbf{53.0} & \underline{26.1} & \textbf{20.7} & \textbf{16.5} & \textbf{12.8} & \underline{5.2} & 7.7 & 9.4 & \underline{3.2} & \underline{3.7}\\
    \bottomrule
  \end{tabular}
  \vspace{10pt}
\end{table*}

On the other hand, we observe that AConE performs closely to the baseline models on answering query types involving negation on dataset NELL-QA, in spite of its consistently good performance on WN18RR-QA. This may be due to two factors: firstly, handling negation is a challenging research question for complex logical query-answering tasks. When it comes to negation queries, all the current models show inferior performance compared to their performance on non-neg queries. Secondly, modeling negation as the complement leads to bias in prediction and high uncertainty in a large number of answers for negation queries. Thus, we leave the task of further improving our model on negation queries for future research. 

\subsection{RQ2: How is the improvement of results related to the capturing of logical patterns?}
To investigate the influence of logical patterns learning on the query-answering model, we provide an ablation analysis considering both the model and data:

\paragraph{\textbf{Model Perspective.}} 
We compared the performance of three AConE's variants: 
AConE (Base), AConE (aperture-multiplicative) and AConE (aperture-additive). They utilize alternative relational rotating transformation strategies for capturing the logical patterns while keeping other modules unchanged. Note that AConE (Base) represents AConE model with any neural relational transformation module. This configuration operates similarly to ConE in terms of its foundational approach. Consequently, we used the results of ConE as a direct representation of "AConE (Base)." This decision was based on their operational similarity, ensuring a fair and coherent presentation of our findings.

\begin{table}[h!]
\centering
\caption{Ablation study of AConE on NELL dataset.}
\resizebox{\hsize}{!}{
\begin{tabular}{llllllllll}
\toprule
 \textbf{Model} & \textbf{1p} & \textbf{2p} & \textbf{3p} & \textbf{2i} & \textbf{3i} & \textbf{pi} & \textbf{ip} & \textbf{2u} & \textbf{up} \\
\midrule
AConE (Base) & 53.1 & 16.1 & 13.9 & 40.0 & 50.8 & 26.3 & 17.5 & 15.3 & 11.3\\
AConE (Ap-Mul) & 51.3 & 16.6 & 13.8 & 38.4 & 48.4 & 18.9 & 19.5 & 14.7 & 12.1\\
AConE (Ap-Add) & \textbf{54.5} & \textbf{17.7} & \textbf{14.4} & \textbf{41.9} & \textbf{53.0} & 26.1 & \textbf{20.7} & \textbf{16.5} & \textbf{12.8}\\
\bottomrule
\end{tabular}}
\label{ablation study}
\end{table}

As Table \ref{ablation study} shows, the performance variation among AConE's variants highlights the impact of relation transformation on the model's ability to capture logical patterns. Moreover, the outperformance of AConE confirms the efficiency of logical patterns in query reasoning tasks.

\begin{figure}
    \centering
    \vspace{-15pt}
    \includegraphics[width=0.8\hsize]{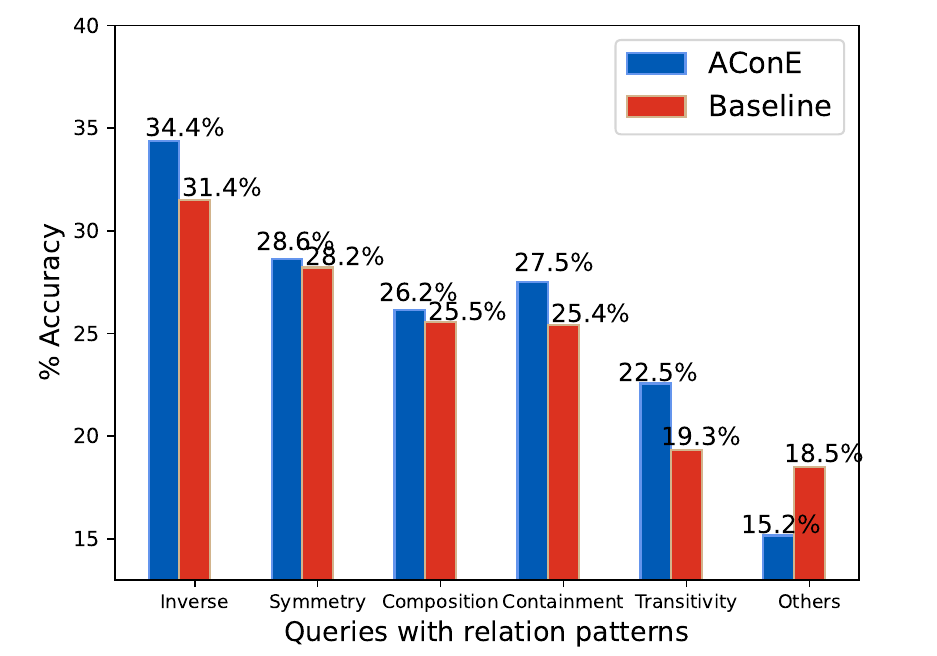}
    \caption{Average performances of AConE and Baseline model over query subgroups with different logical patterns.}
    \vspace{20pt}
    \label{fig:Diff_AVG}
\end{figure}

\paragraph{\textbf{Data Perspective.}} To more effectively examine the specific impact of AConE on queries involving logical patterns, a detailed analysis was conducted on the NELL query answering dataset. We categorize the test dataset into five categories based on the relations involved in the queries. For each query in subgroups \footnote{Note that \emph{Asymmetry} is not included here because such type of relations is missing from the selected rules mined from NELL.} \emph{Inverse}, \emph{Symmetry}, \emph{Composition}, \emph{Containment} and \emph{Transitivity}, there is at least one relation encompassing the corresponding logical pattern. 
Table 7 and Appendix B in the supplementary material \cite{he2024generatingsroiontologiesknowledge} provide more details about these subgroups of queries and elaborate the classification process. The category \emph{Others} corresponds to queries that do not involve any of these logical patterns. Figure \ref{fig:Diff_AVG} shows the average performances of AConE and neural baseline model (ConE) on these subgroups. It is observed that AConE outperforms the neural baseline model on queries that have logical patterns, especially inverse relations. However, AConE does not generalize as well to queries that were not influenced by logical patterns compared to the baseline model. This analysis supports our hypothesis that AConE is better suited for capturing logical patterns in tree-form query answering tasks than purely neural models, with fewer model parameters and a simpler model structure, which is elaborated in section \ref{Analysis on Model Parameters}.

\section{Analysis on Model Parameters}
\label{Analysis on Model Parameters}
The acquisition of logical patterns not only enhances the model performance but also alleviates the reliance on excessively parameterized neural networks in existing methods. Table \ref{Model Parameters} summarizes the number of parameters and the average performance of our method AConE and other baseline models on non-negation queries, which all methods can handle. It shows that AConE has the second-fewest number of parameters among these models, though it achieves much better performances than the baseline models.
\begin{table}
  \caption{Number of parameters in AConE and other baseline models.}
  \label{Model Parameters}
  \centering
  \begin{tabular}{lll}
    \toprule
    \textbf{Model} & \textbf{AVG MRR} & \hspace{5mm} \textbf{Parameters}\\
    \midrule
    BetaE\cite{BetaE} & \hspace{7mm}28.26 &\hspace{7mm}57,574,000\\
    GQE\cite{GQE} & \hspace{7mm}10.52 & \hspace{7mm}52,290,400\\
    ConE\cite{ConE} & \hspace{7mm}31.73 &\hspace{7mm}44,010,401\\
    Q2B\cite{querytobox} & \hspace{7mm}14.06 &\hspace{7mm}26,306,000\\
    \textbf{Ours} & \hspace{7mm}37.64 &\hspace{7mm}36,325,601\\
    \bottomrule
  \end{tabular}
\end{table}


\section{Conclusion and Future Work}
In this work, we investigate the important yet understudied impact of logical pattern inference on the query-answering task. We propose a new embedding method, the multicone embedding, and study its algebraic structure and capability to express $\DLogic$ ontologies. We show some limitations of the method, like breaking the tautology $\exists R.(C \sqcap D) \equiv \exists R.C \sqcup \exists R.D$, the bias favoring the commutativity, and the (general) non-existence of inverse rotations.

Our practical method, {\OurMethod}, is motivated by the multicone algebra, but it has differences. First, multicones consisting of multiple cones are difficult to manage due to the unconstrained number of parameters they require. Hence, we limit the representation of queries to single multicones (i.e., cones). Since this simplification makes the cone intersection no longer closed, we define it as a neural operation returning a single cone. We limit rotations to be aperture-additive because they outperform aperture-multiplicative rotations (see Table~\ref{ablation study}).


Although the experiment datasets had no concept names, multicone embeddings support concept names. We use these datasets to make a fair comparison with the existing methods, which do not support concept names. Since most real datasets include concept names and concept names are a key component of knowledge representation, we plan to extend our research to datasets with concept names. Furthermore, we think that {\OurMethod} can be extended for knowledge graphs including also non-empty RBoxes and TBoxes.


\section*{Ethics Statement}
The authors declare that we have no conflicts of interest. This article does not contain any studies involving business and personal data.

\begin{ack}
  This work was funded by the Deutsche Forschungsgemeinschaft (DFG, German Research Foundation) under the DFG Germany's Excellence Strategy -- EXC 2120/1 -- 390831618, and the DFG Excellence Strategy -- EXC 2075 -- 390740016. We acknowledge the support by the Stuttgart Center for Simulation Science (SimTech). The authors thank the International Max Planck Research School for Intelligent Systems (IMPRS-IS) for supporting Yunjie He. Mojtaba Nayyeri is funded by the BMBF ATLAS project. This work was partially supported by the European Commission funded projects enRichMyData(101070284), GraphMassivizer(101093202), and Dome 4.0 (953163). 
\end{ack}


\bibliography{references}


\appendix

\section{Experimental Details}


\paragraph{Description on Benchmark Datasets}
For a fair comparison with the baseline models, we use the same query structures and logical query datasets from NELL-QA \cite{deeppath} and WN18RR-QA \cite{LinE}, and the open-sourced framework created by \cite{BetaE} for logical query answering tasks. Table \ref{Dataset Statistics} summarizes the statistics of benchmark datasets.

\begin{table}
  \caption{Statistics of benchmark datasets.}
  \label{Dataset Statistics}
  \centering
  \resizebox{\columnwidth}{!}{%
  \begin{tabular}{llllllllll}
    \toprule
    \textbf{Dataset} & \multicolumn{2}{c}{\textbf{Training}} & \multicolumn{2}{c}{\textbf{Validation}}& \multicolumn{2}{c}{\textbf{Testing}}\\
    \midrule & \textbf{1p/2p/3p/2i/3i} & \hspace{2mm}\textbf{Neg} & \hspace{2mm}\textbf{1p} & \textbf{others} & \hspace{2mm}\textbf{1p} & \textbf{others}\\ 
    NELL &  \hspace{5mm}107,980 & 10,798 & 16,927 & 4,000 & 17,034 & 4,000 \\
    WN18RR &  \hspace{5mm}517,545 & 51,750 & 5,202 & 13,000 & 5,356 & 13,000\\
    \bottomrule
  \end{tabular}
  }
\end{table}

\paragraph{\textbf{Hyperparameters and Computational Resources.}}
All of our experiments are implemented in Pytorch \cite{Pytorch} framework and run on four Nvidia A100 GPU cards. For hyperparameters search, we performed a grid search of learning rates in $\{5\times 10^{-5},10^{-4},5\times10^{-4}\}$, the batch size in $\{256, 512, 1024\}$, the negative sample sizes in $\{128,64\}$, the regularization coefficient $\lambda$ in $\{0.02, 0.05, 0.08, 0.1\}$ and the margin $\gamma$ in $\{20, 30, 40, 50\}$. The best hyperparameters are shown in Table \ref{Hyperparameters}.

\begin{table}
  \caption{Hyperparameters found by grid search. d is the embedding dimension, b is the batch size, n is the negative sampling size, $\gamma$ is the margin in loss, l is the learning rate, $\lambda$ is the regularization parameter in the distance function.}
  \label{Hyperparameters}
  \centering
  \begin{tabular}{lllllll}
    \toprule
    \textbf{Dataset} & d & b & n & $\gamma$ & $\hspace{4mm} l$ & $\lambda$\\
    \midrule
    NELL & 800 & 512 & 128 & 20 & $1 \times 10^{-4}$ & 0.02\\
    WN18RR & 800 & 512 & 128 & 20 & $1 \times 10^{-4}$ & 0.02\\
    \bottomrule
  \end{tabular}
\end{table}

\section{Classification Process of Test Queries}
\label{classification of test queries}
This paragraph describes the process for identifying subgroups of test queries with different relation patterns. We first mined rules from the triples-only dataset NELL995 \cite{NELL995} using the rule mining tool AMIE \cite{AMIE}. The rules that are significantly supported by a large amount of data and have body coverages greater than 0.2 are selected. These rules are then categorized into six types: symmetry, inversion, composition, containment, transitivity and other. Each category includes a list of relations that encompass relation patterns. Using the selected relations, the test dataset of the logical query dataset NELL \cite{BetaE} is traversed, and queries are classified into six subgroups: symmetry, inversion, composition, containment, transitivity and other. A query is classified into a subgroup based on the presence of at least one relation from the corresponding relation subgroup. The other subgroup does not include any of these patterns. 

\begin{table*}
  \caption{Statistics on  the percentage of queries that involve various logical patterns in the NELL test dataset}
  \label{subgroup statistics}
  \centering
  \begin{tabular}{lllllllllllllll}
    \toprule
    \textbf{Relation} & \textbf{1p} & \textbf{2p} & \textbf{3p} & \textbf{2i} & \textbf{3i} & \textbf{pi} & \textbf{ip} & \textbf{2u} & \textbf{up} & \textbf{2in} & \textbf{3in} & \textbf{inp} & \textbf{pin} & \textbf{pni}\\
    \midrule
    Symmetry & 20.3	&29.7	&38.6	&25.3	&29.9&	34.4	&31.9&	31.2&	34.6&	27.6&	26.2&	38.6&	37.1&	36.7\\
    Inversion & 26.9&	41.3 & 51.0 &	42.0	&51.9&	51.9&	46.0&	43.2& 43.8 &42.0	&56.1&	51.0	&52.0	&54.3\\
    Composition & 21.6	&35.2&	43.4	&31.5&	41.7	&42.7&	40.1&	35.8&	40.9	&34.7	&42.5	&43.1	&43.0&	43.6\\
    Containment & 37.9 & 51.5 & 63.1 & 50.9 & 59.1 & 54.3 & 48.35 & 48.9 & 49.32 & 48.7 & 50.3 & 60.7 & 60.6 & 58.1\\
    Transitivity & 3.1 & 2.8 & 3.3 & 2.9 & 3.7 & 3.2 & 2.5 & 3.3 & 2.9 & 3.7 & 3.6 & 8.4 & 4.2 & 3.95\\
    \bottomrule
  \end{tabular}
  \vspace{10pt}
\end{table*}

\section{Model Robustness Check}
In this section, we check the model robustness of AConE by running AConE 10 times with different random seeds. The mean performance and standard deviations are reported in Table \ref{EPFOL(sd)}. The slight variances in the results indicate that AConE is robust to the different initializations of model parameters and our reported improvements are not due to randomness.

\section{Intersection Operator}
\label{sec:Intersection}
In this section, we explain the details of two important components of the intersection operator, \textbf{SemanticAverage}($\cdot$) and \textbf{CardMin}($\cdot$) 
\paragraph{SemanticAverage($\cdot$)} is expected to compute the semantic center $\pmb{\theta}_{ax}^{'}$ of the input $\{(\pmb{\theta}_{j,ax}, \pmb{\theta}_{j,ap})\}_{j=1}^{n}$. Specifically, the computation process is provided as:
\begin{equation}
\begin{split}
        [\mathbf{x};\mathbf{y}] &= \sum_{i=1}^{n} [\mathbf{a}_{j}\circ \text{cos}(\pmb{\theta}_{j,ax});\mathbf{a}_{j}\circ \text{sin}(\pmb{\theta}_{j,ax})],\\
    \pmb{\theta}_{ax}^{'} &= \textbf{Arg}(\mathbf{x},\mathbf{y}),
\end{split}
\end{equation}
where cos and sin represent element-wise cosine and sine functions. $\textbf{Arg}$($\cdot$) computes the argument given $\mathbf{x}$ and $\mathbf{y}$. $\mathbf{a}_{j} \in \mathbb{R}^d$ are attention weights such that 
\begin{equation}
    \resizebox{0.95\hsize}{!}{%
        $[\mathbf{a}_{j}]_{k} = \frac{\text{exp}([\textbf{MLP}([\pmb{\theta}_{j,ax} - \pmb{\theta}_{j,ap/2};\pmb{\theta}_{j,ax} + \pmb{\theta}_{j,ap/2}])]_{k})}{\sum_{m=1}^{n}\text{exp}([\textbf{MLP}([\pmb{\theta}_{m,ax} - \pmb{\theta}_{m,ap/2};\pmb{\theta}_{m,ax} + \pmb{\theta}_{m,ap/2}])]_{k})}, $%
        }
\end{equation}
where \textbf{MLP} : $\mathbb{R}^{2d} \rightarrow \mathbb{R}^{d}$ is a multi-layer perceptron network, [$\cdot$;$\cdot$] is the concatenation of two vectors. 

\paragraph{CardMin($\cdot$)} predicts the aperture $\pmb{\theta}_{ap}^{'}$ of the intersection set such that $[\pmb{\theta}_{ap}^{'}]_{i}$ should be no larger than any $\pmb{\theta}_{j,ap}^{i}$, since the intersection set is the subset of all input entity sets.
\begin{equation}
    \resizebox{0.98\hsize}{!}{%
    $[\theta_{ap}^{'}]_{i} = \text{min}\{\theta_{1,ap}^{i},\dots,\theta_{n,ap}^{i}\}\cdot\sigma([\textbf{DeepSets}(\{(\pmb{\theta}_{j,ax},\pmb{\theta}_{j,ap})\}_{j=1}^{n})]_{i})$%
    }
\end{equation}
where \textbf{DeepSets}$(\{(\pmb{\theta}_{j,ax}, \pmb{\theta}_{j,ap})\}_{j=1}^{n})$ \citep{deepsets} is given by
\begin{equation*}
    \resizebox{0.95\hsize}{!}{%
    $\textbf{MLP}(\frac{1}{n}\sum_{j=1}^{n}\textbf{MLP}([\pmb{\theta}_{j,ax} - \pmb{\theta}_{j,ap/2};\pmb{\theta}_{j,ax} + \pmb{\theta}_{j,ap/2}])).$%
    }
\end{equation*}

\begin{table*}[t]
\caption{AConE's MRR mean values and standard variances ($\%$) on answering tree-form ($\exists$, $\land$, $\lor, \neg$) queries}
\label{EPFOL(sd)}
\centering
\begin{tabular}{lllllllllllllll}
\hline
\textbf{Dataset} & \textbf{1p} & \textbf{2p} & \textbf{3p} & \textbf{2i} & \textbf{3i} & \textbf{pi} & \textbf{ip} & \textbf{2u} & \textbf{up} & \textbf{2in} & \textbf{3in} & \textbf{inp} & \textbf{pin} & \textbf{pni}\\
\hline
$\multirow{2}{5em}{NELL}$ & 54.5$\pm$ & 17.7$\pm$ & 14.4$\pm$ & 41.9$\pm$ & 53.0$\pm$ & 26.1$\pm$ & 20.7$\pm$ & 16.5$\pm$ & 12.8$\pm$ & 5.2$\pm$ & 7.7$\pm$ & 9.4$\pm$ & 3.2$\pm$ & 3.7$\pm$\\
& 0.097 & 0.152 & 0.192 & 0.130 & 0.038 & 0.084 & 0.146 & 0.112 & 0.137& 0.012 & 0.079 & 0.154 & 0.012 & 0.097\\
\hline
\end{tabular}
\end{table*}

\end{document}